\documentclass[runningheads]{llncs}
\usepackage{graphicx}
\usepackage{times}
\usepackage{epsfig}
\usepackage{amsmath}
\usepackage{amssymb}
\usepackage{booktabs}
\usepackage{wrapfig}
\begin{document}
\title{Automated Seed Quality Testing System using GAN \& Active Learning}
%
%
\author{Sandeep Nagar\inst{1} \and
Prateek Pani\inst{1} \and
Raj Nair\inst{2} \and
Girish Varma\inst{1}}
\authorrunning{Sandeep Nagar et al.}

\institute{Machine Learning Lab, International Institute of Information Technology, Hyderabad, India 
\and AdTech Corp, Hyderabad, India \\
\email{sandeep.nagar, prateek.pani@research.iiit.ac.in, raj@adtechcorp.in, girish.varma@iiit.ac.in}}
\maketitle              
\begin{abstract}

Quality assessment of agricultural produce is a crucial step in minimizing food stock wastage. However, this is currently done manually and often requires expert supervision, especially in smaller seeds like corn. We propose a novel computer vision-based system for automating this process. We build a novel seed image acquisition setup, which captures both the top and bottom views. Dataset collection for this problem has challenges of data annotation costs/time and class imbalance. We address these challenges by i.) using a Conditional Generative Adversarial Network (CGAN) to generate real-looking images for the classes with lesser images and ii.) annotate a large dataset with minimal expert human intervention by using a Batch Active Learning (BAL) based annotation tool. We benchmark different image classification models on the dataset obtained. We are able to get accuracies of up to 91.6\% for testing the physical purity of seed samples.

\end{abstract}

\section{Introduction}
Quality checking is an essential step to ensuring food grain supplies. It ensures that stocks with different compositions of defective grains are not mixed and can be processed, packaged, and sold for appropriate uses, from high quality, economy packaging to animal feeds. However, quality checking for small grains and seeds like corn, rice is a tedious task done mainly through experts by visual inspection. Manual inspections are not scalable because of the human resources required, inconsistency between inspectors, and slower processing pipeline. An automated approach to seed/grain quality testing can solve many of these problems, leading to better usage and distribution of food stocks. 
\begin{figure}[ht]
    \centering
    \includegraphics[width=.95\linewidth]{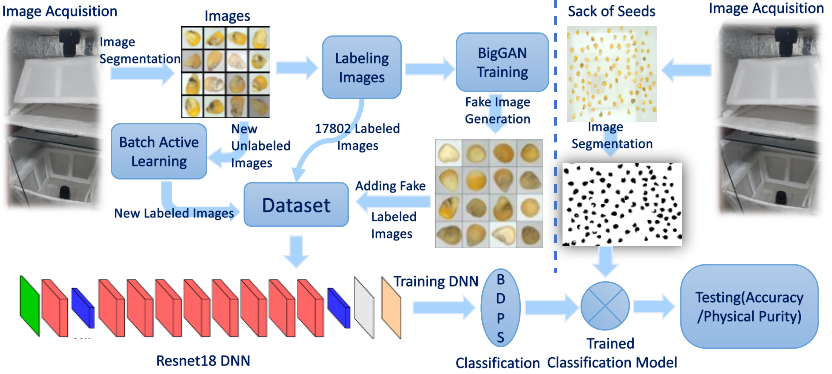}
    \vspace{-1em}
    \caption{Overview of our proposed system.}
    \label{fig:my_label}
    \vspace{-2em}
\end{figure}

With the advent of Deep Neural Networks (DNNs), data-driven models have become increasingly adept at image classification/detection tasks. DNN models have been used in literature for seed quality testing problems \cite{seeds_corn,seeds_corn2,cornseed}. However, there are some significant impediments to their widespread adoption. DNNs require large-scale datasets for giving high accuracies, which matches human inspectors. However, creating large-scale, good-quality datasets is challenging. Annotating seed data sets with defects requires experts with specialized knowledge. Also, a considerable imbalance of seeds can occur in a sample with a specific defect, which results in small datasets highly skewed to non-defective classes, with minimal images in some defective categories. Another main problem is that the defective part of the seed might not be visible from the top view.

We propose addressing class imbalance and annotation costs using machine learning techniques such as generative methods and BAL. We use active learning to select the most informative images from unlabeled data to be labelled by experts, leading to lesser annotation. We use generative methods, expressly conditional generative adversarial networks, to generative images of each class to address the class imbalance problem. We also develop a novel hardware design, which examines the seeds from the top and bottom, leading to better classification accuracy.  

Active learning uses to select an initial batch of images to be annotated. After a round of annotation, the next batch is determined based on the uncertainty of classification remaining conditioned on the initial batch of annotation. Batch Active Learning also ensures that the images within the same batch are diverse compared to basic Active Learning methods. We build an annotation tool, which shows the next batch obtained by Active Learning. It also incorporates UI elements like suggestions for labels from the partially trained model on images labelled so far, which leads to short annotation. Using this tool, we build a dataset of 26K corn seed images, classified into pure seeds and three other defective classes (broken, silkcut, and discoloured).

However, the dataset created is highly imbalanced, with images of pure seeds being $40\%$ while some of the defective classes being as low as $9\%$ with $4$ classes in total. We use a Conditional Generative Adversarial Network (CGAN) to address the class imbalance problem. A CGAN (BigGAN) trained on the labelled base dataset conditioned on the labels to generate good quality seed images. Then the CGAN is used to generate images that are indistinguishable from real images for each class. Hence this results in a more balanced dataset. Finally, an image classifier DNN is trained on this dataset (with the fake images added), leading to better accuracy of 80\%.

\paragraph{Main Contributions.}
We build a computer vision-based system for large-scale automated quality checking of corn seeds. 
\begin{enumerate}
    \item We give a better hardware design, which takes images from the top and bottom to inspect defective/pure seeds (see Section \ref{sec:hard-prim-data}).
    \item We build an annotation tool using Batch Active Learning and specific UI elements to accelerate the annotation process (see Section \ref{sec:active-learning}).
    \item We use Conditional Generative Adversarial Networks to generate fake images of each class, leading to a larger and more balanced training dataset (see Section \ref{sec:cgan}).
    \item We build a dataset of 26K corn seed images labeled as pure, broken, silkcut, and discolored (see Section \ref{subsec:dataset}).
    \item We train an image classifier on the dataset with generated and real images, leading to improved accuracy (see Section \ref{sec:results}).
\end{enumerate}

\section{Related Works}

\paragraph{ML for Seed Quality Testing and Agriculture.}

Machine vision for precision agriculture has attracted research interest in recent years \cite{ml_in_agri,TIAN20201,rice,cornseed}. Plant health monitoring approaches are addressed, including weed, insect, and disease detection \cite{ml_in_agri}. With the success of DNNs, different approaches have been proposed to tackle problems of corn seed classification \cite{cornseed,seeds_corn,seeds_corn2}. Fine-grained objects (seeds) are visually similar by a rough glimpse, and details can correctly recognize them in discriminative local regions.

\paragraph{Generative Methods for Class Imbalance.}
Generative models can not only be used to generate images \cite{gan_for_class}, but adversarial learning showed good potential in restoring balance in imbalanced datasets \cite{gan_plant}. Generative models can generate samples that are difficult for the object detector to classify. Creating a balanced dataset is a problem because the availability of one type of sample (seed) with defects or impurity compared to others is not always the same. While creating a new dataset, the imbalance of the instance, and we applied the fake image generation to overcome this. We use a generative model for the Image-to-image translation with conditional adversarial networks\cite{gan_l}.

\paragraph{Batch Active Learning for Fast Annotation.} Manual data annotation can be very slow and costly needs expertise for the same. There is no single standard format when it comes to image label/annotation. In our dataset, images contain only seeds, and each image is labelled one class out of four classes. Labelling the fine-grained image is challenging due to the significant intraclass variance and slight inter-class variance to recognize hundreds of sub-categories belonging to the same basic-level category.  The aim of Active Learning (AL) is to discover the dependence of some variable ($y$) on an input variable ($x$) \cite{BALD}. We use the Batch Active Learning (BatchBALD) \cite{BatchBALD} to label the images with the help of a small no. of manually labelled images.

\section{Approach}
We approach the problem of seed quality testing by first building a camera setup and preprocessing pipeline to obtain individual seed images from a sample which are then labelled (see Section \ref{sec:hard-prim-data}). Then, since the data is highly imbalanced and to minimize the expert labelling effort, we propose two methods: i.) use Active learning-based UI tool to aid the creation of a larger dataset with the least effort from the expert human intervention (see Section \ref{sec:active-learning}) ii.) using Conditional Generative Adversarial Networks (CGAN) to generate images to solve the class imbalance problem (see Section \ref{sec:cgan}).
\subsection{Hardware Setup and Primary Dataset Creation.} \label{sec:hard-prim-data}
Our seed quality testing approach makes use of a camera setup shown in Figure.\ref{fig:setup} below. In this setup, we use two cameras, one on the top and one bottom. We place a bunch of seeds in the middle on transparent glass, take a picture one from each camera and save it to the computer connected to the camera. To block the other side of the transparent glass, we use a thin white film. While capturing the top view of the seeds, we put the thin film below the glass, which works as a white background, similar to the bottom view. Thus, the top camera gives the picture of a top view of each seed, and similarly, the bottom camera capture the bottom picture of seeds to train the classification model. We use the top and bottom images of seeds as individual input images, which can get two independent predictions, increasing the accuracy.

\begin{wrapfigure}{l}{0.4\textwidth}
    \vspace{-2em}
    \includegraphics[width=0.99\linewidth]{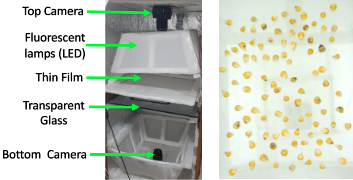} 
    \vspace{-1em}
    \caption{Image capturing setup and sample image.}
    \label{fig:setup}
    \vspace{-2em}
\end{wrapfigure}
For seed classification, individual corns in the captured image should be accurately segmented first. During placing the seeds on the transparent glass, we mind the gap between the neighbour seeds to avoid the overlapping of seeds segmentation \cite{seg} of the image done by the Watershed method because this transform decomposes an image completely and assigns each pixel to a region or a watershed, and image segmentation can be accomplished simultaneously. After the segmentation, the expert's labelling of each seed image is done for the 17802, considering the top and bottom two different seed images. To classify new seeds after the training model, we take a picture of a sack of seeds and use the segmentation to detect the location \cite{seg} of each seed and give it as the input to the model classification.

\subsection{Batch Active Learning (BAL) for Fast Annotation.} \label{sec:active-learning}
Data efficiency is a crucial problem in Deep Learning. Active learning, a sub-field in Machine learning, is centred around attaining data efficiency. To avoid the tedious data labelling of a large dataset, in Active Learning, we iteratively query the most informative points from a set of unlabelled points. However, in practical Active learning, rather than acquiring single points, we query a batch of points from an unlabelled set of most informative and diverse points. The question is which subset of points of the unlabelled set should be added to the training set so that the model would learn the fastest when trained on this updated training set than picking any other subset of the unlabelled points. 

Moreover, since we can label multiple images in a single screen shown as a grid to the annotator (see Fig \ref{fig:anno_ui}), there is a possibility of sampling points belonging to the same class of the underlying distribution. In such a scenario, we will have a model being accurate in one class but not in all other classes. To avoid this situation, we query the least confidence, and the queried set should be diverse. Batch Active Learning (BAL) was proposed to solve such problems, which we use to build an annotation tool.



\subsection{Conditional Image Generation to Balance Dataset}\label{sec:cgan}

\begin{wrapfigure}{l}{0.5\textwidth}
\vspace{-2em}
 \centering
 \includegraphics[width=0.9\linewidth]{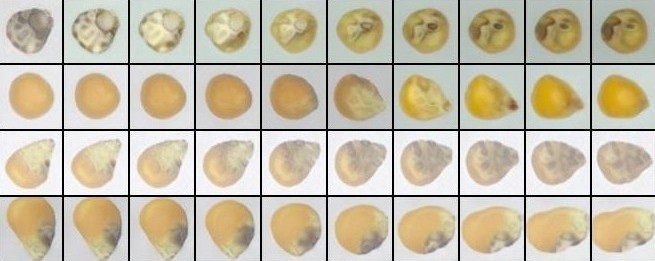} 
  \vspace{-1em}
 \caption{Example of interpolation two GAN generated images. A gradual linear transformation from one seeds to another using latent vector(z) interpolation.}
\label{fig:inter}
\vspace{-1.5em}
\end{wrapfigure}
Generative Adversarial Networks (GANs) are one of the state-of-the-art approaches for image generation. Existing works primarily aim at synthesizing data of high visual quality \cite{gan_l,gan_high}. We use BigGAN \cite{gan_l}, as it gives good quality images in similar datasets. Training generative adversarial networks (GANs) using too little data typically leads to discriminator (D) over-fitting \cite{limitdata}. GANs training is dynamic and sensitive to nearly every aspect of its setup (from optimization parameters to model architecture). \textbf{Image Interpolation:} Exploring the structure of the \emph{latent space} $(z)$ for a GANs model is interesting for the problem domain. It helps develop an intuition for what has been learned by the generating model (see Figure \ref{fig:inter}). We use image interpolation to evaluate whether the inverted codes are semantically meaningful. We interpolate one type of seed to other classes in a large diversity. As we can see in Figure \ref{fig:inter} left to right, how smoothly the interpolation works, which validates that the GAN has learned a good latent representation for images in the dataset.

\section{Results}\label{sec:results}

\subsection{Dataset \& Experimental Details} 
\label{subsec:dataset}
Our primary dataset contains four classes: \emph{broken}, \emph{discolored}, \emph{pure}, and \emph{silkcut} (B, D, P, S), having different instances for each class (Imbalance). Images are taken for both sides of the corn, \emph{top-view}($8901$) and \emph{bottom-view}($8901$), as explained above in the hardware setup section-\ref{sec:hard-prim-data}. The primary dataset is highly imbalanced. To analyze the class confusion for seeds, we explore the confusion matrix in the result section\ref{sec:results}. Instances of each class and their \% in the primary dataset are as follows: \emph{broken} $5670$ $(32\%)$, \emph{discolored} $3114$ $(17.4\%)$, \emph{pure} $7267 (40.8\%)$, \emph{silkcut} $1751 (09.8\%)$. We use two methods to add more images into the primary dataset: i.) adding GAN generated images to balance the dataset and ii.) adding more captured images labeled using the Batch Active Learning method.

In case of adding fake images to balance the dataset, we split the primary dataset into two parts train and validation in a 70:30 ratio, ensuring that each class has the same \% of instances on each set. The train set is used to train the BigGAN model, and after adding fake images generated by BigGAN into the train set, this new train set is used to train the classification model. Finally, the Validation set is used for testing the classification model only. 

We generated fake images as follows: \emph{broken} $2937$, \emph{discolored} $5823$, \emph{pure} $2937$, \emph{silkcut} $5823$ instances and added them into the train set to balance the data set. In the case of adding newly captured images labelled using the \emph{Batch Active Learning} method after image segmentation, new $9000$ labelled images are added into the \emph{primary} dataset. This new dataset contains 26,802 images split into Train and Validation set $80:20$, respectively. The train set is used to train the classification model, and the Validation part is used to test the classification model. 

We train different Convolutional Neural Networks (CNN) models on this dataset in Section \ref{subsec:classification}. First, we use transfer learning; the models are initialized with weights learned from ImageNet Classification \cite{imnetclass}. Then, we trained the DNNs on the train set of the primary dataset to fine-tune the model and compare the validation accuracy for a different model (see table\ref{tab:freq}). Since Resnet$18$ has the highest accuracy, we trained it on three different datasets: i.) the primary dataset, ii.) with fake images generated using BigGAN, and iii.) after adding the newly annotated images labelled by the Batch Activation Learning method.

\subsection{Batch Active Learning (BAL)}\label{subsec:batch}
\begin{figure}[ht!]
    \centering
    \vspace{-1em}
    \includegraphics[width=0.8\linewidth]{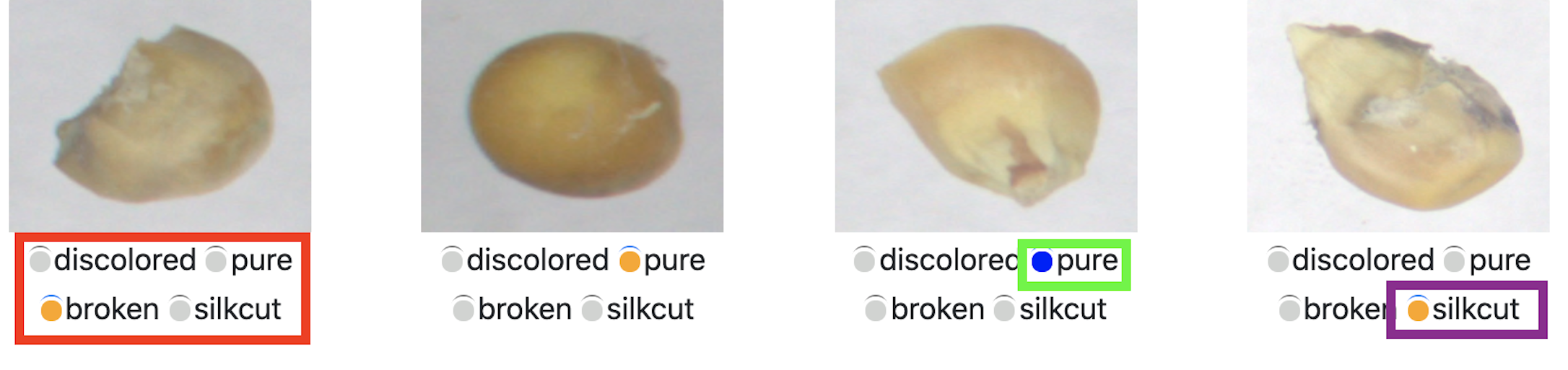}
        \vspace{-1.5em}
    \caption{The user interface of the annotation tool. A single-screen shows a batch of images.  If the suggestion is wrong, the user can relabel it. After a few active learning cycles, the model suggestions become more accurate, resulting in less effort from the annotator. Purple box: label suggested by the model.}
    \label{fig:anno_ui}
    \vspace{-1em}
\end{figure}

We start with the corn seed \emph{primary} dataset with 17,802 images and do a 90\%-10\% train-validation split to train the BAL model. We also have an unlabelled dataset of corn seeds of 26,777 to label using BAL. Specifically, we use the BatchBALD \cite{BatchBALD} method.  An active learning cycle involves retraining a model on the data annotated so far. During training, we use early stopping to avoid over-fitting. The used acquisition function is based on model uncertainty and ensures that the queried images are diverse in the predicted distribution models uncertainty entropy. Next, pick  5000 images that have the highest entropy (most informative points). To ensure diversity, perform k-means clustering on these 5000 images until stabilization to find the 1000 cluster centres. After that, find the points in the dataset closest to each of these 1000 centres. This spits out 1000 images queried by the above acquisition function along with their predicted labels. We then render these images and their predicted labels via an annotation tool that we have built for human annotation (see Figure.\ref{fig:anno_ui}). Next step, the training set is updated by accumulating these images and the labels that the annotator makes. A reinitialized model is then trained on the updated training set, and the cycle continues.

\begin{figure}[ht!]
  \centering
  \includegraphics[width=0.8\linewidth]{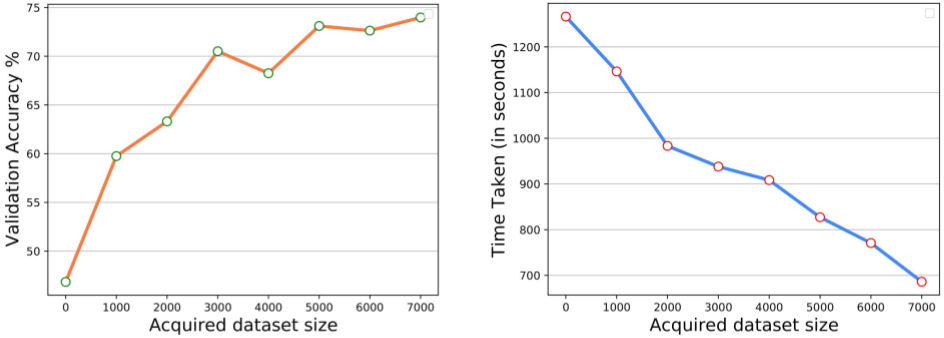}
  \vspace{-1.5em}
  \caption{(Left) Over 8 AL cycles shows an increase from 46.83 to 73.97\% accuracy. (Right) Time taken to annotate shows a drastic decrease in annotation time over 8 AL cycles.}
  \label{fig:batch_learning}
  \vspace{-2em}
\end{figure}
We record the validation accuracy for every cycle and find that with labelling only 9000 images of the 26,777 original unlabelled set, there is a significant increase in the validation accuracy from 46.83\% to 73.97\% which is shown in Figure \ref{fig:batch_learning}.
Moreover, the annotation time is taken to annotate the first queried batch of 1000 images fell drastically from 1266.42 seconds to 682.02 seconds (which is the time taken to annotate the $8$th queried batch) as reflected in Figure \ref{fig:batch_learning}, thereby decreasing the annotation cost and efficiency significantly. After the $9$th cycle, the total labelled instances in the dataset are 26802 after adding new labelled images. 
\vspace*{-\baselineskip}

\subsection{Conditional Image Generation (CGAN)}\label{subsec:CGAN}
\begin{figure}[ht!]
\vspace{-2em}
  \centering
  \includegraphics[width=\linewidth]{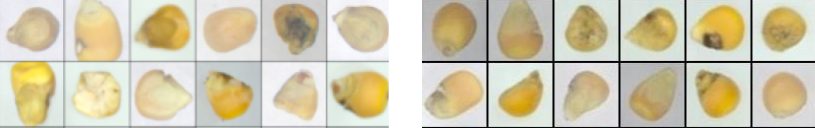}
  \vspace{-2em}
  \caption{Sample images from primary dataset (left) \& ones generated using BigGAN (right).}
  \vspace{-1.5em}
  \label{fig:sample_img}
\end{figure}

We trained the conditional GANs to generate the fake images to add to the dataset and reduce the class instances' imbalance. To train BigGAN, the \emph{primary} dataset is split into two sets, train and validation $70:30$, respectively. The train set is used to train the BigGAN and, after adding the fake images, to train the classification model. The validation set is used only to test the classification model. We trained the Generator (G) and Discriminator (D) of the BigGAN for 250 epochs, and after training, we passed a random noise and label to the G to generate the fake images. Hyper-parameters  used are as follows: input image resolution$=256\times256\times3$, learning rate $=2e^{-4}$, batch size $=16$, dimension of the latent vector space $\text{dim}(z)=128$. BigGAN is trained in alternate phases for D, G and we ensure that each input batch contains an equal no. of images from each class. Figure \ref{fig:sample_img} shows and compares that the image generated is indistinguishable from real images.
\vspace{-2em}
\subsection{Classification}\label{subsec:classification}
We have validated the effectiveness of Active Learning in Section \ref{subsec:batch} and analyzed the quality of images generated by the Conditional GAN in Section \ref{subsec:CGAN}. Here we discuss the accuracy of the classification model trained on the dataset obtained.

\begin{table}[t]
\begin{minipage}{.33\textwidth}\centering
\vspace{1em}
  \begin{tabular}{lll}
    \toprule
    Model & $\qquad$ & Acc. \%   \\
    \midrule
    \textbf{resnet18} &  &  \textbf{71}\\
    squeezenet &  &  71 \\
    resnet50 &   & 70 \\
    mobilenet &  &  68 \\
    wideResnet50 &  &  69 \\
  \bottomrule
  \end{tabular}
  \vspace{0.5em}
  \caption{Validation accuracy for different CNN models on \emph{primary} dataset. }
  \label{tab:freq}
\end{minipage}
\vspace{-3em}
\hfill
\begin{minipage}{.62\textwidth}
\centering
  \begin{tabular}{ccccccl}
    \toprule
    \# fake images &$\qquad$ & Acc.\% & $\qquad$ & Phys. Purity \%  \\
    \midrule
    Zero && 71.00 && 80.58\\
    20K && \textbf{79.24} && \textbf{88.25} \\ 
    40K && 78.23 && 87.68 \\
    100K && 78.88 && 87.24\\
  \bottomrule
  \end{tabular}
  \vspace{0.5em}
  \caption{Validation accuracy comparison before and after adding the fake images to dataset(for resnet18 DNN) solving imbalance. \emph{Accuracy:} four class classification accuracy. \emph{Physical Purity}: Pure vs Impure(Two class classification) }
  \label{tab:fake}
\end{minipage}
\end{table}

\begin{figure*}[ht!]
\vspace{1em}
  \centering
  \includegraphics[width=0.9\textwidth]{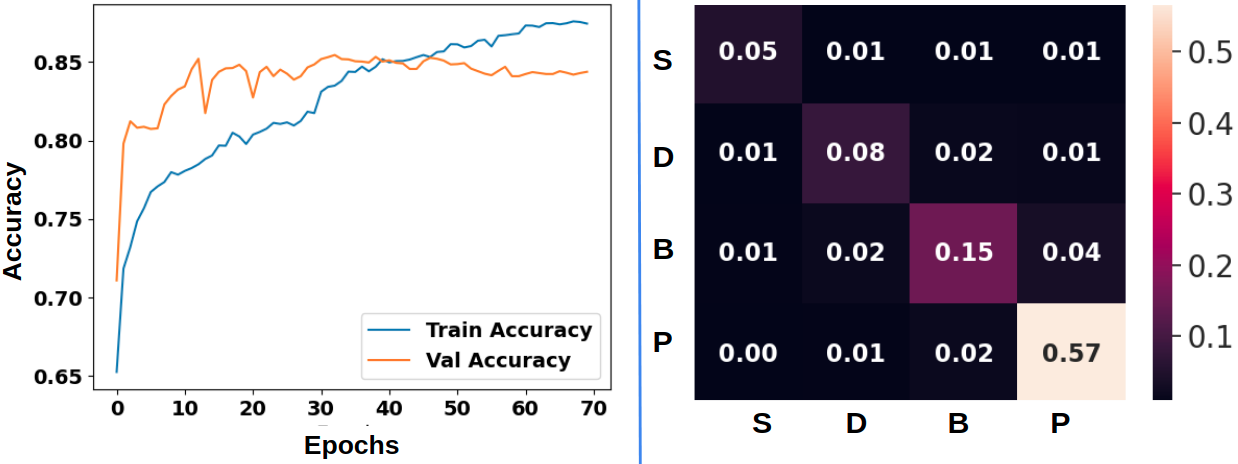}
  \vspace{-1em}
  \caption{(Left) Train and Validation accuracy of the ResNet18 using pre-trained model. (Right) Confusion Matrix: heat map of validation images (4946) after training the Resnet18 on train set of new dataset (26802). Ground Truth: rows, Predicted: columns. B: broken, D: discolored, S: silkcut, P: pure.}
  \label{fig:heatmap}
\vspace{-2.5em}
\end{figure*}
The highest validation accuracy of the primary dataset is $71.02\%$ (see Table \ref{tab:freq}) among different deep learning models. In the case of adding fake images into the train set of the primary dataset to balance the dataset, the validation accuracy after training the classification model on the new train set increases the form $71.02\%$ to $79.24\%$ (see Table \ref{tab:fake}), thereby validating the approach of using CGAN to solve the class imbalance problem. The graph in Figure \ref{fig:heatmap} plots the training and validation accuracy for the Resnet 18 on the dataset after adding new images labeled by the BAL. This further improves the accuracy to $85.24\%$ and we also get the \emph{physical purity} (\emph{pure} vs \emph{impure} classification) accuracy to be $91.62\%$. The classwise accuracies are as follows: \emph{broken} $71.20\%$, \emph{discolored} $69.08\%$, \emph{pure} $94.94\%$, \emph{silkcut} $75.82\%$.

Some images in the dataset have high-class ambiguity. To analyze, we used the confusion matrix given in Figure \ref{fig:heatmap} for the validation set. Each matrix entry gives the \% of images of specific ground truth (rows) and a specific predicted class (columns). As can be seen from Figure \ref{fig:heatmap}, the classes \emph{pure} and \emph{broken} are most confusing followed by \emph{discolored} and \emph{broken}.

\section{Conclusions \& Discussion}
We propose a novel computer vision-based automated system that can be used for corn seed quality testing. A novel image acquisition setup is used so that two different viewpoints are obtained for every seed. Furthermore, we decrease the human intervention required for labelling by building a BAL based UI tool. We also address the class imbalance problem by using Conditional GANs (BigGAN) to generate more images of classes with a small dataset. We believe similar approaches can be used for quality testing of various seeds and vegetables and can decrease wastage and human intervention.

\end{document}